# Beyond Task Success: Measuring Workflow Fidelity in LLM-Based Agentic Payment Systems

Donghao HUANG[1,2], Joon Kiat CHUA[1], and Zhaoxia WANG[1]

[1] School of Computing and Information Systems, Singapore Management University, 80 Stamford Rd, Singapore 178902, Singapore

[2] Research and Development, Mastercard, 4250 Fairfax Dr, Arlington, VA 22203, USA

**Abstract.** LLM-based multi-agent systems are increasingly deployed for payment workflows, yet prevailing metrics—Task Success Rate (TSR) and Agent Handoff F1-Score (HF1)—capture only final outcomes or unordered routing decisions. We introduce the *Agentic Success Rate* (ASR), a trajectory-fidelity metric that compares observed and expected agent execution sequences at the *transition* (bigram) level, decomposing performance into Transition Recall (completeness) and Transition Precision (efficiency). Applied to the Hierarchical Multi-Agent System for Payments (HMASP) across 18 LLMs and 90,000 task instances (5 repeats per model–scenario pair), ASR reveals that 10 of 18 models systematically skip a confirmation checkpoint during payment checkout—invisible to both TSR and HF1—while 8 models enforce the checkpoint perfectly. Notably, `GPT-4.1` exhibits hidden workflow shortcuts despite achieving perfect TSR and HF1, while `GPT-5.2` achieves perfect ASR. Prompt refinements and deterministic routing guards guided by ASR diagnostics yield substantial TSR improvements, with gains up to +93.8 pp for previously struggling models, demonstrating that trajectory-level evaluation is essential in regulated domains.

## 1 Introduction

The emergence of autonomous LLM-based agents has fueled rapid adoption of *agentic payments*—using LLM agents to automate end-to-end payment workflows [16]. Major payment networks have launched agentic capabilities: Mastercard's Agent Pay [12] and Visa's Intelligent Commerce [17], with projections reaching $1.7 trillion by 2030 [16]. In research, frameworks like HMASP [5] have demonstrated the feasibility of hierarchical multi-agent systems for payments.

Despite these deployments, current evaluation frameworks remain inadequate. HMASP, for example, relies on Task Success Rate (TSR) and Agent Handoff F1-Score (HF1). However, TSR captures only final outcomes, and HF1 evaluates routing in a bag-of-edges fashion, ignoring temporal order. Neither captures *workflow fidelity*—the agent's adherence to the expected sequence of steps. In payment systems, auditability is critical [14], and the sequential path to the result is as important as the outcome itself [4, 3].

Drawing on process mining conformance checking [1, 2], we propose the **Agentic Success Rate (ASR)**, a trajectory-level metric that compares agent-to-agent *transitions* (consecutive handoff pairs) between observed and expected workflows. Our contributions are: (1) ASR as a multi-agent payment metric; (2) evaluation of 18 LLMs on 90,000 instances; (3) discovery of workflow shortcuts invisible to TSR/HF1; and (4) demonstration that ASR-guided engineering yields dramatic TSR improvements for previously struggling models.

## 2 Related Work

**LLM-Based MAS for Payments.** Multi-agent systems have been applied to software engineering [8], trading [18], and investment [20]. In payments, prior work addressed process automation [7] and fraud detection [6] but not end-to-end agentic workflows. HMASP [5] filled this gap with a hierarchical agent framework.
**Evaluating Agentic Systems.** Standard evaluation focuses on task completion [10, 9], but binary metrics obscure behavioral nuances. AgentBoard [11] introduced Progress Rate; WorFEval [15] uses subsequence matching for workflow planning; $\tau$-bench [19] proposed pass$^k$ for reliability; and Chen et al. [4] argued standard benchmarks underestimate financial risks. ASR builds on this line of work [13], grounding trajectory evaluation in process mining conformance checking [1, 2].

## 3 Methodology

### 3.1 Agentic Success Rate (ASR)

ASR compares the full ordered trajectory against the expected one at the *transition* level, drawing on the conformance checking paradigm from process mining [1, 2]. Each consecutive pair of agents in a trajectory is a transition (bigram); for example, the trajectory $[A, B, C]$ yields the transition multiset $\{(A,B),\ (B,C)\}$. Given expected trajectory $E = [e_1, \ldots, e_n]$ and observed trajectory $O = [o_1, \ldots, o_m]$, let $B_E$ and $B_O$ denote their respective transition multisets:

$$\text{TR} = \frac{|B_E \cap B_O|}{|B_E|}, \quad \text{TP} = \frac{|B_E \cap B_O|}{|B_O|}, \quad \text{ASR} = F_1(\text{TR}, \text{TP}) \qquad (1)$$

An ASR of 100% indicates perfect transition alignment. Low Transition Recall (TR) signals skipped handoffs ("model moves" in process mining terminology);

low Transition Precision (TP) signals redundant ones ("log moves"). Because ASR operates on transitions rather than individual agent visits or unordered edge sets, it directly measures whether each agent handed off control to the correct next agent—precisely the property needed for compliance auditing—and generalizes to arbitrary agentic architectures.

### 3.2 Evaluation Setup

**Models.** We evaluated 18 models spanning both open-weight and proprietary LLMs. From the original HMASP study [5], we include Qwen3 (8B, 14B, 32B), Qwen2.5 (7B, 14B, 32B), Mistral-Small-3.2 (24B), Magistral (24B), Llama 3.1 (8B, 70B), and GPT-4.1. Additionally, we evaluated seven models not included in the original study: GPT-5.2, GPT-OSS (20B, 120B), and Gemma4 (E2B, E4B, 26B, 31B). Proprietary models (GPT-4.1, GPT-5.2) are accessed via the official OpenAI API, while all open-weight models are run locally using Ollama v0.20.0 on an Apple M5 Max with 128 GB unified memory and 4-bit quantization (MXFP4 for GPT-OSS models; Q4_K_M for all others).
**Dataset and Repeat Protocol.** We use the HMASP evaluation dataset of 1,000 data points across four scenarios (250 each): **T1** (card registration; e.g., *"Add a new card"*), **T2** (card retrieval; e.g., *"Show my saved cards"*), **T3** (payment processing; e.g., *"Complete my purchase"*), and **T4** (irrelevant input rejection; e.g., *"What's the weather today?"*). Each model–scenario pair is evaluated 5 times independently, yielding 90,000 total instances. We report the mean and standard deviation across the 5 repeats.
**Metrics.** We report three metrics, the first two from the original HMASP evaluation [5]: **(1) Task Success Rate (TSR):** the proportion of runs completing the correct workflow with correct information saved or retrieved; **(2) Agent Handoff F1-Score (HF1):** the F1-score comparing actual vs. expected agent-to-agent handoffs; and **(3) Agentic Success Rate (ASR):** trajectory-level workflow fidelity as defined above.
**System Engineering.** We improve the HMASP system through two complementary techniques, both guided by ASR diagnostics. **(i) Prompt engineering:** agent prompts are refined iteratively to better specify roles, expected handoffs, and task boundaries. **(ii) Deterministic routing guards:** a *cart-items guard* in the payment supervisor suppresses the payment-review path when `cart_items` is empty, and *card-view pattern matching* in the CPA provides keyword-based fallback routing for ambiguous card intents.

## 4 Results and Discussion

### 4.1 Effectiveness of System Engineering

Table 1 compares TSR between HMASP [5] and our system for the three most-improved models (ranked by average $\Delta$ across all tasks). The gains are striking: Llama3.1:8b improves by an average of +67.9 pp, Magistral:24b by +54.2 pp, and Llama3.1:70b by +33.5 pp—confirming that failures were addressable through engineering rather than model replacement.

**Table 1.** HMASP vs. Ours: TSR (%) for the three most-improved models. H = HMASP, O = Ours (mean over 5 repeats), $\Delta$ = O - H. Bold: $\geq$20 pp gains.

| Model | T1 | | | T2 | | | T3 | | | T4 | | |
|---|---|---|---|---|---|---|---|---|---|---|---|---|
| | H | O | $\Delta$ | H | O | $\Delta$ | H | O | $\Delta$ | H | O | $\Delta$ |
| Llama3.1:8b | 6.0 | 99.8 | **+93.8** | 10.4 | 98.8 | **+88.4** | 4.0 | 93.4 | **+89.4** | 100 | 100 | +0.0 |
| Magistral:24b | 44.0 | 90.0 | **+46.0** | 18.4 | 97.7 | **+79.3** | 3.2 | 94.7 | **+91.5** | 100 | 100 | +0.0 |
| Llama3.1:70b | 66.0 | 100 | **+34.0** | 74.0 | 100 | **+26.0** | 26.0 | 98.9 | **+72.9** | 98.8 | 100 | +1.2 |

### 4.2 ASR Reveals Hidden Workflow Deviations

Table 2 reports T3 metrics (mean ± std over 5 repeats) and average across T1–T4 for all 18 models, sorted by average ASR. Eight models achieve a perfect 100% on all three T3 metrics: all four Gemma4 variants, both GPT-OSS models, `MSmall3.2:24b`, and `GPT-5.2`. Among the 10 models that exhibit a T3 ASR gap, the range spans from 0.01% (`Qwen2.5:32b`) to 5.45% (`Qwen2.5:7b`). The highlighted cells are the sharpest examples: `Qwen2.5:32b`, `GPT-4.1`, `Qwen3:32b`, and `Qwen3:8b` achieve T3 TSR = HF1 = 100% yet ASR < 100%, exposing skipped confirmation checkpoints invisible to conventional metrics.

**Table 2.** T3 (Payment) results (%, mean ± std over 5 repeats) and average across T1–T4 for all 18 models. Sorted by Avg ASR desc. Yellow: TSR = HF1 = 100% but ASR < 100%.

| Model | T3 (Payment) | | | Avg (T1–T4) | | |
|---|---|---|---|---|---|---|
| | TSR | HF1 | ASR | TSR | HF1 | ASR |
| *8 models*[†] | 100 | 100 | 100 | 100 | 100 | 100 |
| Qwen2.5:32b | 100 | 100 | 99.99±0.02 | 100 | 100 | 100.00±0.00 |
| GPT-4.1 | 100 | 100 | 99.96±0.03 | 100 | 100 | 99.99±0.01 |
| Qwen3:32b | 100 | 100 | 99.97±0.04 | 99.98±0.04 | 99.99±0.02 | 99.97±0.05 |
| Qwen3:14b | 99.68±0.33 | 99.84±0.17 | 99.64±0.31 | 99.92±0.08 | 99.96±0.04 | 99.91±0.08 |
| Qwen3:8b | 100 | 100 | 99.97±0.02 | 99.90±0.07 | 99.95±0.04 | 99.89±0.08 |
| Llama3.1:70b | 98.88±1.21 | 98.78±1.09 | 98.81±1.17 | 99.72±0.30 | 99.69±0.27 | 99.70±0.29 |
| Qwen2.5:14b | 99.76±0.22 | 99.80±0.20 | 99.56±0.16 | 99.32±0.39 | 99.64±0.22 | 99.56±0.23 |
| Llama3.1:8b | 93.44±0.92 | 96.05±0.72 | 93.31±0.97 | 98.02±0.41 | 98.81±0.28 | 98.03±0.36 |
| Magistral:24b | 94.72±1.18 | 97.05±0.63 | 93.88±1.04 | 95.60±1.01 | 97.65±0.54 | 95.52±0.95 |
| Qwen2.5:7b | 53.28±2.05 | 64.16±2.28 | 47.83±1.96 | 77.54±0.91 | 85.12±0.79 | 77.09±0.71 |

[†]Gemma4:e4b/26b/31b, GPT-OSS:20b/120b, MSmall3.2:24b, GPT-5.2.

The root cause is consistent across all 10 models exhibiting a T3 ASR gap. The expected T3 trajectory has 11 agent hops (10 transitions), including an intermediate return to the CPA for explicit user *confirmation* before payment processing. For direct payment phrases (e.g., "Process my payment"), models infer unambiguous intent and skip this checkpoint, executing 9 hops (8 transitions). Of 10 expected transitions, 8 appear: TR = 8/10 = 80%, TP = 8/8 = 100%, ASR = $F_1(80\%, 100\%) = 88.9\%$ per affected sample. Crucially, every single hidden deviation across all affected models follows this *identical* 9-step shortcut—there is exactly one deviant trajectory pattern, reflecting a systematic interaction between input phrasing and model reasoning rather than stochastic variation.

This has regulatory significance: PCI-DSS [14] requires auditable workflows, and bypassing confirmation checkpoints creates gaps in the audit trail even when outcomes are correct.

## 5 Conclusion

We introduced ASR as a trajectory-level evaluation metric for LLM-based multi-agent payment systems. Across 18 LLMs and 90,000 instances, ASR exposes a systematic procedural deviation: 10 of 18 models skip a CPA-mediated confirmation checkpoint during payment checkout (T3 ASR gaps of 0.01–5.45%) despite near-perfect TSR and HF1, while 8 models enforce the checkpoint perfectly. Notably, `GPT-4.1` exhibits hidden workflow shortcuts despite achieving perfect TSR and HF1, while `GPT-5.2` achieves perfect ASR. ASR-guided system engineering yields dramatic improvements for previously struggling models (up to +93.8 pp). As payment networks deploy agentic capabilities at scale [12, 17], ASR provides the visibility needed to ensure procedural fidelity—essential where compliance, auditability, and liability attribution are at stake. Future work includes validating ASR in production systems, weighted variants for security-critical transitions, architectural enforcement of the T3 checkpoint, and cross-architecture evaluations [17].